\documentclass{article}
\usepackage{spconf,amsmath,graphicx,hyperref}
\usepackage{booktabs}
\usepackage{adjustbox}
\usepackage{multirow}
\usepackage{amsfonts}
\usepackage{array}
\usepackage{xcolor}
\usepackage[table]{xcolor}
\usepackage{subcaption} 
\newcommand{\xhdr}[1]{\vspace{1mm}\noindent{{\bf #1.}}}
\usepackage{cleveref}

\title{Dynamic Multi-Expert Projectors with Stabilized Routing for Multilingual Speech Recognition}
\name{
Isha Pandey${^1}$ \qquad
Ashish Mittal${^{1,2}}$\qquad 
Vartul Bahuguna${^1}$ \qquad 
Ganesh Ramakrishnan${^1}$
}

\address{
\emph{IIT Bombay${^1}$, IBM Research${^2}$} \\
\small{ \small{\{iishapandey, vartul, ganesh\}@cse.iitb.ac.in}, arakeshk@in.ibm.com
}
}

\begin{document}
%
\maketitle
\begin{abstract}
Recent advances in LLM-based ASR connect frozen speech encoders with Large Language Models (LLMs) via lightweight projectors. While effective in monolingual settings, a single projector struggles to capture the diverse acoustic-to-semantic mappings required for multilingual ASR. To address this, we propose \textbf{SMEAR-MoE}, a stabilized Mixture-of-Experts projector that ensures dense gradient flow to all experts, preventing expert collapse while enabling cross-lingual sharing. We systematically compare monolithic, static multi-projector, and dynamic MoE designs across four Indic languages (Hindi, Marathi, Tamil, Telugu). Our SMEAR-MoE achieves strong performance, delivering upto a 7.6\% relative WER reduction over the single-projector baseline, while maintaining comparable runtime efficiency. Analysis of expert routing further shows linguistically meaningful specialization, with related languages sharing experts. These results demonstrate that stable multi-expert projectors are key to scalable and robust multilingual ASR. We have released our code at \footnote{https://github.com/iishapandey/SMEAR-MoE-ASR.git}
\end{abstract}
\section{Introduction}
\label{sec:intro}

Automatic Speech Recognition (ASR) systems have become a cornerstone technology for voice-driven applications such as virtual assistants, transcription services, and accessibility tools. With the rapid emergence of large-scale foundation models, ASR has increasingly benefited from advances in representation learning and cross-modal integration. A particularly promising direction is to connect a frozen ASR encoder with a Large Language Model (LLM), enabling the system to leverage the linguistic and world knowledge of LLMs while requiring minimal task-specific training \cite{ma2024embarrassingly, saon2025granite}. This paradigm is especially appealing for resource-constrained languages, where high-quality ASR datasets are scarce but LLMs already encode substantial cross-lingual knowledge.

Despite their effectiveness, current LLM-based ASR systems face a major bottleneck in the \textit{projector} that bridges the speech encoder and the LLM. In multilingual settings, a single monolithic projector struggles to capture the diverse acoustic-to-semantic mappings of typologically distinct languages (e.g., Indo-Aryan vs. Dravidian), forcing representational compromises. In our experiments, we observe that \textit{language-specific projectors} improve per-language accuracy but hinder cross-lingual sharing, while \textit{mixture-of-experts (MoE)} designs enable specialization yet often suffer from instability and \textit{expert} collapse, where only a few experts dominate due to poor gradient flow.

We propose \textbf{SMEAR-MoE}, a stabilized Mixture-of-Experts projector that ensures dense gradient flow, combining expert specialization with cross-lingual sharing while avoiding collapse in conventional MoEs. On four mid-resource Indic languages (Hindi, Marathi, Tamil, Telugu), SMEAR-MoE achieves significant relative WER reduction over the single projector baseline and further outperforms static ensembles and standard MoEs. Beyond accuracy, it prevents expert collapse and enables related languages (e.g., Hindi and Marathi) to share experts, yielding interpretable specialization that aligns with linguistic families. This interpretability underscores SMEAR-MoE’s potential as a scalable and robust solution for multilingual ASR with LLMs.

In summary, we make the following contributions: 1) We analyze the \textit{projector bottleneck} in multilingual ASR and benchmark language-specific and expert-based alternatives. 2) We propose \textbf{SMEAR-MoE}, a stabilized Mixture-of-Experts projector that mitigates expert collapse by ensuring dense gradient flow. 3) On four mid-resource Indic languages, SMEAR-MoE achieves up to 7.6\% WER reduction while maintaining the efficiency of lightweight projectors. 4) Routing analysis shows that SMEAR-MoE learns \textit{linguistically meaningful expert sharing}, enabling interpretable and robust multilingual ASR.

\section{Related Work}

Recent advances in \textit{LLM-based ASR} connect frozen speech encoders with Large Language Models (LLMs) through lightweight projectors~\cite{ma2024embarrassingly,saon2025granite,yu2024connecting}. Architectures such as SLAM-ASR show that even simple adapters can bridge acoustic and textual spaces, enabling efficient training and strong zero-shot capabilities. To improve multilingual performance, several works introduce \textit{adapter-based methods}. MMS~\cite{pratap2024scaling} add small language-specific modules to a common backbone, while parameter-efficient tuning with LoRA~\cite{hu2022lora} assigns lightweight adapters per language. These approaches enhance accuracy, especially for low- and mid-resource languages, but limit cross-lingual sharing and scale poorly as the number of languages grows. Recent work such as MOSA~\cite{li2025mosa} highlights this trade-off, proposing mixtures of adapters to balance shared and language-specific capacity.

A complementary line of work employs \textit{Mixture-of-Experts (MoE)} layers, which expand capacity by routing inputs to specialized experts while keeping computation sparse~\cite{shazeer2017outrageously,cappellazzo2025scaling}. MOSA~\cite{li2025mosa} shows MoE adapters outperform monolithic projectors, while HDMoLE~\cite{mu2025hdmole} uses hierarchical routing for accent robustness. However, conventional MoEs often suffer from instability and expert collapse, especially in low- and mid-resource settings. Large-scale systems such as Whisper~\cite{whisper} underscore the potential of cross-lingual sharing, but projector stability remains an open challenge. Our work addresses this with \textbf{SMEAR-MoE}, a stabilized MoE projector~\cite{muqeethsoft} that ensures dense gradient flow, prevents collapse, and enables interpretable expert sharing across related languages.

\begin{figure}[t] 
    \centering
    \includegraphics[width=\columnwidth]{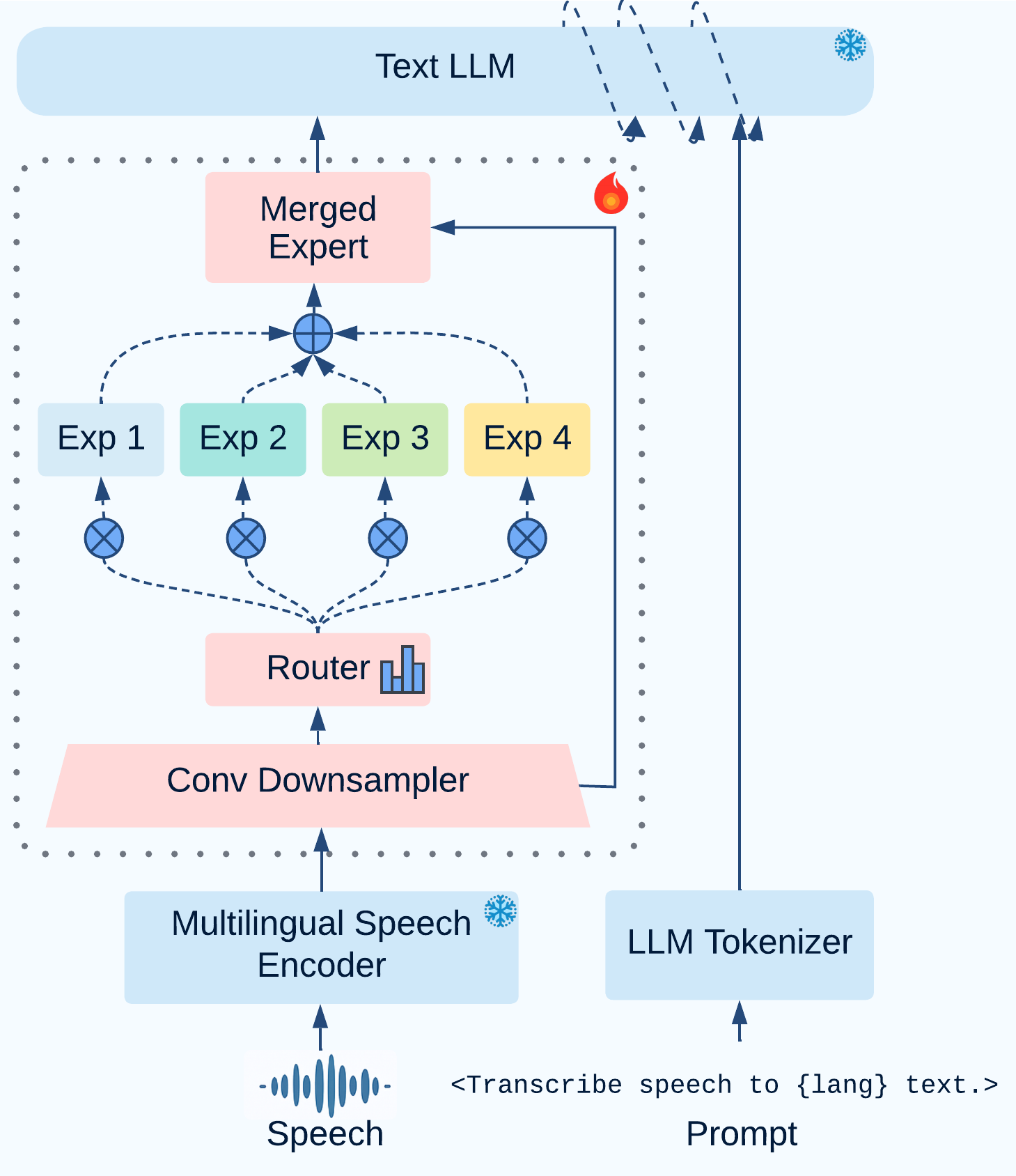} 
    \caption{Illustration of \textbf{SMEAR-MoE}. Unlike hard-gated MoEs that route to a few experts, SMEAR merges all experts into a single virtual expert, applied to the downsampled features. This ensures dense gradient flow, stable training, and prevents expert collapse.}

    \label{fig:indic_slam}
\end{figure}

\section{Methodology}
\label{sec:methodology}

Our work builds upon the SLAM-ASR framework~\cite{ma2024embarrassingly}, which connects a frozen speech encoder to a frozen LLM via a lightweight, trainable projector. We redesign this projector for multilingual ASR by systematically progressing from monolithic to multi-expert architectures. Formally, given an input speech signal $X_S$, the encoder produces a sequence of hidden states:
\begin{equation}
    H_S = \text{Encoder}(X_S).
\end{equation}

\subsection{Monolithic Projector}
The projector maps the sequence $H_S$ into the LLM embedding space, typically after downsampling. We first consider a convolutional–MLP design:
\begin{equation}
    E_S = \text{MLP}\big(\text{ReLU}(\text{Conv1D}(H_S))\big).
    \label{eq:mono_proj}
\end{equation}
While effective in monolingual settings, extending this design to multilingual ASR introduces a bottleneck: a single projector must align heterogeneous acoustic patterns with text embeddings. Increasing its depth initially yields small gains but soon degrades performance due to negative interference across languages. This motivates a shift toward multi-expert architectures.

\subsection{Static Multi-Projector Architectures}

We next evaluated static multi-projector variants:projectors:
\begin{itemize}
    \item \textbf{Language-Specific Projectors:} each language $m$ uses its own dedicated expert $P_m$, i.e., $E_S = P_m(H_S)$ from equation \ref{eq:mono_proj}. This prevents sharing and performs poorly under limited data.
    \item \textbf{Tied Projectors:} languages are grouped by family (e.g., Indo-Aryan, Dravidian), and outputs of experts in group $\mathcal{G}_m$ are averaged:
    \begin{equation}
        E_S = \frac{1}{|\mathcal{G}_m|} \sum_{j \in \mathcal{G}_m} P_j(H_S).
    \end{equation}
    \item \textbf{Dense Ensemble:} all experts contribute equally:
    \begin{equation}
        E_S = \frac{1}{M} \sum_{m=1}^{M} P_m(H_S).
    \end{equation}
\end{itemize}
The dense ensemble performs best, confirming the benefit of shared representations. However, its static averaging and high computational cost motivate a more dynamic, efficient design.

\subsection{Dynamic Projector via Mixture-of-Experts}
To enable dynamic specialization, we adopt a Mixture-of-Experts (MoE) design. We decouple the projector into:
\begin{enumerate}
    \item a shared \textbf{convolutional downsampler} $D(\cdot)$ that reduces sequence length:
    \begin{equation}
        Z_S = D(H_S) = \text{Conv1D}\big(\text{ReLU}(\text{Conv1D}(H_S))\big),
    \end{equation}
    \item a set of lightweight \textbf{expert MLPs} $\{P_m\}_{m=1}^M$ operating on $Z_S$.
\end{enumerate}
A gating network produces token-level probabilities over experts:
\begin{equation}
    G = \text{Softmax}(Z_S W_g),
\end{equation}
where $G \in \mathbb{R}^{T \times M}$ and $T$ is the sequence length.  

We implement two standard top-$k$ MoE strategies:
\begin{itemize}
    \item \textbf{Utterance-Level MoE:}
    Averaging token-level gates yields $\bar{g}$, which determines Top-$k$ expert routing:
    \begin{equation}
    \bar{g} = \frac{1}{T} \sum_{t=1}^{T} G_t \in \mathbb{R}^M \;;\quad
    E_S = \sum_{j \in \text{Top-}k(\bar{g})} \bar{g}_j \, P_j(Z_S).
    \label{eg:utter_level}
    \end{equation}

    \item \textbf{Token-Level MoE:} apply top-$k$ gating per token $t$, then recombine:
    \begin{equation}
        E_{S,t} = \sum_{j \in \text{Top-}k(G_t)} G_{t,j} \, P_j(Z_{S,t}).
    \end{equation}
\end{itemize}
These methods suffer from sparse gradient flow: only selected experts receive updates, leading to instability and expert under-utilization.

\subsection{Stabilized MoE Routing with SMEAR-MoE}
To overcome this, we adopt SMEAR~\cite{muqeethsoft}, which constructs a differentiable \textit{virtual expert} by merging all expert parameters according to gating weights $\bar{g}$ (see \ref{eg:utter_level}). Given expert parameters $(W_m, b_m)$, we compute:
\begin{equation}
    \bar{W} = \sum_{m=1}^{M} \bar{g}_m W_m, \quad
    \bar{b} = \sum_{m=1}^{M} \bar{g}_m b_m.
\end{equation}
The virtual expert $(\bar{W}, \bar{b})$ is applied to $Z_S$ and , producing $E_S$. Unlike hard routing, this ensures every expert receives a dense gradient signal proportional to $\bar{g}_m$, eliminating expert collapse and yielding stable training even in data-constrained multilingual settings. 

\begin{table*}[!hbt]
\centering
\small
\resizebox{\textwidth}{!}{%
\begin{tabular}{ll|cc|cc|cc|cc|cc|cc|cc|cc}
\toprule
\rowcolor[HTML]{FFFFFF} 
\multicolumn{1}{c}{\cellcolor[HTML]{FFFFFF}\textbf{Lang}} & \textbf{Datasets} & \multicolumn{2}{|c}{\cellcolor[HTML]{FFFFFF}\textbf{\begin{tabular}[c]{@{}c@{}}Whisper large-v3\end{tabular}}} & \multicolumn{2}{|c}{\cellcolor[HTML]{FFFFFF}\textbf{Single Projector}} & \multicolumn{2}{|c}{\cellcolor[HTML]{FFFFFF}\textbf{\begin{tabular}[c]{@{}c@{}}Lang-Specific \\ Projector\end{tabular}}} & \multicolumn{2}{|c}{\cellcolor[HTML]{FFFFFF}\textbf{Tied Projector}} & \multicolumn{2}{|c}{\cellcolor[HTML]{FFFFFF}\textbf{Dense Ensemble}} & \multicolumn{2}{|c}{\cellcolor[HTML]{FFFFFF}\textbf{\begin{tabular}[c]{@{}c@{}}Utterance-Level \\ MoE\end{tabular}}} & \multicolumn{2}{|c}{\cellcolor[HTML]{FFFFFF}\textbf{\begin{tabular}[c]{@{}c@{}}Token-Level \\ MoE\end{tabular}}} & \multicolumn{2}{|c}{\cellcolor[HTML]{FFFFFF}\textbf{SMEAR MoE}} \\
\rowcolor[HTML]{FFFFFF} 
 & \multicolumn{1}{l}{\cellcolor[HTML]{FFFFFF}} & \multicolumn{2}{|c}{\cellcolor[HTML]{FFFFFF}} & \multicolumn{2}{|c}{\cellcolor[HTML]{FFFFFF}\textbf{18.16 M}} & \multicolumn{2}{|c}{\cellcolor[HTML]{FFFFFF}\textbf{72.64 M}} & \multicolumn{2}{|c}{\cellcolor[HTML]{FFFFFF}\textbf{72.64 M}} & \multicolumn{2}{|c}{\cellcolor[HTML]{FFFFFF}\textbf{72.64 M}} & \multicolumn{2}{|c}{\cellcolor[HTML]{FFFFFF}\textbf{52.98 M}} & \multicolumn{2}{|c}{\cellcolor[HTML]{FFFFFF}\textbf{52.98 M}} & \multicolumn{2}{|c}{\cellcolor[HTML]{FFFFFF}\textbf{52.98 M}} \\
 \hline
\rowcolor[HTML]{FFFFFF} 
 & \multicolumn{1}{l}{\cellcolor[HTML]{FFFFFF}} & \multicolumn{1}{|c}{\cellcolor[HTML]{FFFFFF}\textbf{CER}} & \multicolumn{1}{c}{\cellcolor[HTML]{FFFFFF}\textbf{WER}} & \multicolumn{1}{|c}{\cellcolor[HTML]{FFFFFF}\textbf{CER}} & \multicolumn{1}{c}{\cellcolor[HTML]{FFFFFF}\textbf{WER}} & \multicolumn{1}{|c}{\cellcolor[HTML]{FFFFFF}\textbf{CER}} & \multicolumn{1}{c}{\cellcolor[HTML]{FFFFFF}\textbf{WER}} & \multicolumn{1}{|c}{\cellcolor[HTML]{FFFFFF}\textbf{CER}} & \multicolumn{1}{c}{\cellcolor[HTML]{FFFFFF}\textbf{WER}} & \multicolumn{1}{|c}{\cellcolor[HTML]{FFFFFF}\textbf{CER}} & \multicolumn{1}{c}{\cellcolor[HTML]{FFFFFF}\textbf{WER}} & \multicolumn{1}{|c}{\cellcolor[HTML]{FFFFFF}\textbf{CER}} & \multicolumn{1}{c}{\cellcolor[HTML]{FFFFFF}\textbf{WER}} & \multicolumn{1}{|c}{\cellcolor[HTML]{FFFFFF}\textbf{CER}} & \multicolumn{1}{c}{\cellcolor[HTML]{FFFFFF}\textbf{WER}} & \multicolumn{1}{|c}{\cellcolor[HTML]{FFFFFF}\textbf{CER}} & \multicolumn{1}{c}{\cellcolor[HTML]{FFFFFF}\textbf{WER}} \\
 \hline
\rowcolor[HTML]{FFFFFF} 
\cellcolor[HTML]{FFFFFF} & fleurs & 13.3 & 32.5 & 9.2 & 17.8 & 20.3 & 27.7 & 11.4 & 19.9 & \cellcolor[HTML]{CCCCFF}8.2 & \cellcolor{green!30}16.6 & 10.5 & 19.5 & 10.2 & 18.4 & 9.4 & 17.4 \\
\rowcolor[HTML]{FFFFFF} 
\cellcolor[HTML]{FFFFFF} & indictts & 8.6 & 28.8 & \cellcolor[HTML]{CCCCFF}7.9 & \cellcolor{green!30}26.6 & 12.1 & 31.2 & 8.2 & 27.4 & 8.1 & 27.1 & 8.1 & 27.0 & 8.9 & 27.4 & \cellcolor[HTML]{CCCCFF}7.9 & 27.0 \\
\rowcolor[HTML]{FFFFFF} 
\cellcolor[HTML]{FFFFFF} & kathbath & 12.0 & 32.5 & 5.8 & 12.3 & 9.3 & 14.9 & 5.3 & 12.0 & \cellcolor[HTML]{CCCCFF}5.2 & 11.8 & 5.9 & 12.7 & 6.1 & 12.9 & 5.4 & \cellcolor{green!30}11.4 \\
\rowcolor[HTML]{FFFFFF} 
\cellcolor[HTML]{FFFFFF} & mucs &  12.2 & 32.6 & 8.0 & 19.4 & 10.6 & 21.0 & 9.1 & 20.1 & 8.0 & 19.1 & 8.3 & 19.4 & 10.3 & 20.8 & \cellcolor[HTML]{CCCCFF}7.9 & \cellcolor{green!30}18.4 \\
\hline 
\rowcolor[HTML]{FFFFFF} 
 \multirow{-5}{*}{\cellcolor[HTML]{FFFFFF}Hindi} & \textbf{Average} & \textbf{11.5} & \textbf{31.6} & \textbf{7.7} & \textbf{19.0} & \textbf{13.1} & \textbf{23.7} & \textbf{8.5} & \textbf{19.9} & \cellcolor[HTML]{CCCCFF}\textbf{7.4} & \textbf{18.7} & \textbf{8.2} & \textbf{19.7} & \textbf{8.9} & \textbf{19.9} & \textbf{7.7} & \cellcolor{green!30}\textbf{18.6} \\
\hline \hline
\rowcolor[HTML]{FFFFFF} 
\cellcolor[HTML]{FFFFFF} & fleurs & \cellcolor[HTML]{FFFFFF}24.9 & \cellcolor[HTML]{FFFFFF}80.7 & 9.6 & 27.5 & 14.2 & 31.4 & 12.2 & 29.1 & 10.5 & 28.0 & 11.2 & 29.5 & 11.1 & 28.6 & \cellcolor[HTML]{CCCCFF}9.4 & \cellcolor{green!30}25.9 \\
\rowcolor[HTML]{FFFFFF} 
\cellcolor[HTML]{FFFFFF} & indictts & \cellcolor[HTML]{FFFFFF}15.9 & \cellcolor[HTML]{FFFFFF}67.7 & 13.1 & 48.5 & 6.9 & 31.2 & 6.9 & 31.7 & 6.6 & 30.2 & 6.7 & 31.3 & 6.8 & 30.7 & \cellcolor[HTML]{CCCCFF}6.5 & \cellcolor{green!30}30.0 \\
\rowcolor[HTML]{FFFFFF} 
\cellcolor[HTML]{FFFFFF} & kathbath & \cellcolor[HTML]{FFFFFF}25.5 & \cellcolor[HTML]{FFFFFF}84.9 & 8.8 & 25.0 & 11.6 & 28.0 & 9.8 & 25.9 & 9.4 & 25.2 & 9.4 & 25.8 & 10.7 & 26.2 & \cellcolor[HTML]{CCCCFF}8.2 & \cellcolor{green!30}23.2 \\
\rowcolor[HTML]{FFFFFF} 
\cellcolor[HTML]{FFFFFF} & mucs & \cellcolor[HTML]{FFFFFF}24.8 & \cellcolor[HTML]{FFFFFF}67.4 & 5.1 & 26.8 & 35.8 & 52.1 & 5.4 & \cellcolor{green!30}26.0 & 5.3 & 27.2 & 5.6 & 26.9 & 6 & 28.6 & \cellcolor[HTML]{CCCCFF}4.9 & 26.1 \\
\rowcolor[HTML]{FFFFFF} 
\hline
\multirow{-5}{*}{\cellcolor[HTML]{FFFFFF}Marathi} & \textbf{Average} & \textbf{22.8} & \textbf{75.2} & \textbf{9.2} & \textbf{32.0} & \textbf{17.1} & \textbf{35.7} & \textbf{8.6} & \textbf{28.2} & \textbf{8.0} & \textbf{27.7} & \textbf{8.2} & \textbf{28.4} & \textbf{8.7} & \textbf{28.5} & \cellcolor[HTML]{CCCCFF}\textbf{7.3} & \cellcolor{green!30}\textbf{26.3} \\
\hline \hline
\rowcolor[HTML]{FFFFFF} 
\cellcolor[HTML]{FFFFFF} & fleurs & 14.2 & 52.8 & 14.0 & 33.8 & 17.6 & 42.7 & 12.8 & 32.7 & 12.7 & 31.8 & 12.4 & 31.7 & 12.9 & 32.9 & \cellcolor[HTML]{CCCCFF}11.8 & \cellcolor{green!30}30.5 \\
\rowcolor[HTML]{FFFFFF} 
\cellcolor[HTML]{FFFFFF} & indictts & 12.4 & 54.9 & 7.1 & 47.3 & 16.0 & 59.4 & 10.6 & 50.2 & 6.8 & 44.4 & 7.3 & 45.7 & 8.1 & 49.9 & \cellcolor[HTML]{CCCCFF}6.6 & \cellcolor{green!30}44.2 \\
\rowcolor[HTML]{FFFFFF} 
\cellcolor[HTML]{FFFFFF} & kathbath & 13 & 58.6 & 6.3 & 28.7 & 12.6 & 45.0 & 6.1 & 28.4 & 6.1 & 28.7 & 6.4 & 29.4 & 7.3 & 30.3 & \cellcolor[HTML]{CCCCFF}5.5 & \cellcolor{green!30}27.1 \\
\rowcolor[HTML]{FFFFFF} 
\cellcolor[HTML]{FFFFFF} & mucs & 13.8 & 55.8 & 8.3 & 31.0 & 14.7 & 44.2 & 8.9 & 31.7 & 7.8 & 30.2 & 9.1 & 31.6 & 9.9 & 33.5 & \cellcolor[HTML]{CCCCFF}7.2 & \cellcolor{green!30}29.0 \\
\hline
\rowcolor[HTML]{FFFFFF} 
\multirow{-5}{*}{\cellcolor[HTML]{FFFFFF}Tamil} & \textbf{Average} & \textbf{13.4} & \textbf{55.5} & \textbf{8.9} & \textbf{35.2} & \textbf{15.2} & \textbf{47.8} & \textbf{9.6} & \textbf{35.8} & \textbf{8.4} & \textbf{33.8} & \textbf{8.8} & \textbf{34.6} & \textbf{9.6} & \textbf{36.7} & \cellcolor[HTML]{CCCCFF}\textbf{7.8} & \cellcolor{green!30}\textbf{32.7} \\
\hline \hline
\rowcolor[HTML]{FFFFFF} 
\cellcolor[HTML]{FFFFFF} & fleurs & 56.9 & 116.4 & 11.5 & 29.6 & 11.0 & 29.8 & 10.9 & 29.2 & \cellcolor[HTML]{CCCCFF}10.3 & \cellcolor{green!30}27.9 & 12.3 & 31.2 & 11.3 & 30.5 & 10.4 & 28.6 \\
\rowcolor[HTML]{FFFFFF} 
\cellcolor[HTML]{FFFFFF} & indictts & 65.6 & 90.9 & 8.2 & \cellcolor{green!30}48.3 & 10.4 & 50.5 & 11.5 & 51.4 & 8.5 & 50.9 & 8.8 & 50.4 & 9.8 & 52.3 & \cellcolor[HTML]{CCCCFF}8.1 & 48.9 \\
\rowcolor[HTML]{FFFFFF} 
\cellcolor[HTML]{FFFFFF} & kathbath & 53.0 & 106.5 & 5.6 & 27.1 & 6.3 & 27.6 & 5.7 & 26.9 & 5.4 & 26.6 & 6.5 & 27.8 & 6.7 & 28.1 & \cellcolor[HTML]{CCCCFF}5.2 & \cellcolor{green!30}25.5 \\
\rowcolor[HTML]{FFFFFF} 
\cellcolor[HTML]{FFFFFF} & mucs & 68.7 & 138.1 & 11.4 & \cellcolor{green!30}34.8 & 12.1 & 36.4 & 13.7 & 37.8 & 11.5 & 35.4 & \cellcolor[HTML]{CCCCFF}11.3 & 35.4 & 13.5 & 37.9 & 11.9 & 35.3 \\
\rowcolor[HTML]{FFFFFF} 
\hline
\multirow{-5}{*}{\cellcolor[HTML]{FFFFFF}Telugu} & \textbf{Average} & \textbf{61.1} & \textbf{113.0} & \textbf{9.2} & \textbf{35.0} & \textbf{10.0} & \textbf{36.1} & \textbf{10.5} & \textbf{36.3} & \cellcolor[HTML]{CCCCFF}\textbf{8.9} & \textbf{35.2} & \textbf{9.7} & \textbf{36.2} & \textbf{10.3} & \textbf{37.2} & \cellcolor[HTML]{CCCCFF}\textbf{8.9} & \cellcolor{green!30}\textbf{34.6} \\
\hline \hline

\rowcolor[HTML]{FFF2CC} 
\multicolumn{2}{l|}{\textit{\textbf{Overall Average}}} & \textbf{27.2} & \textbf{68.8} & \textbf{8.7} & \textbf{30.3} & \textbf{13.8} & \textbf{35.8} & \textbf{9.3} & \textbf{30.0} & \textbf{8.2} & \textbf{28.8} & \textbf{8.7} & \textbf{29.7} & \textbf{9.4} & \textbf{30.6} & \cellcolor[HTML]{CCCCFF}\textbf{7.9} & \cellcolor{green!30}\textbf{28.0} \\

\bottomrule
\end{tabular}
}
\caption{Comparative performance (WER/CER) of ASR models on four Indian language datasets, benchmarking \textbf{SMEAR-MoE} against baseline Whisper large-v3, SLAM-ASR, and architectures with Static and Dynamic MoE Projectors.}
\vspace{-0.2cm}
\label{tab:result_table}
\end{table*}

\section{Experimental Setup}

\xhdr{Model Specifications}
We use a frozen Whisper large-v3 multilingual encoder~\cite{whisper} and a frozen Gemma-2-9B LLM~\cite{mesnard2024gemma}, with training restricted to the projector. The baseline is a monolithic Conv1D projector ($\sim$18M parameters). A static multi-projector ensemble scales this to $\sim$72M, while our MoE design employs a shared convolutional downsampler ($\sim$13M) and four lightweight MLP experts ($\sim$9M each), totaling $\sim$52M. This is more efficient than the dense ensemble and enables dynamic expert specialization. Gemma was chosen as the text backbone since it was state-of-the-art for Indic languages at the time.

\xhdr{Data Specifications}
We train on IndicVoices and IndicSUPERB~\cite{javed2024indicvoices,javed2023indicsuperb}, sampling $\sim$250 hours per language (Hindi, Marathi, Tamil, Telugu) to avoid confounding effects. Evaluation uses the VISTAR Benchmark~\cite{bhogale2023vistaar}, including Kathbath, MUCS~\cite{diwan2021mucs}, IndicTTS~\cite{kumar2023towards}, Fleurs~\cite{conneau2023fleurs}, covering diverse speaking styles from studio read speech to conversational and crowdsourced audio.

\xhdr{Training and Inference} We follow the SLAM-ASR format~\cite{ma2024embarrassingly}, conditioning inputs on language-specific prompts. Models are trained for up to 60k steps using AdamW (lr $1\!\times\!10^{-3}$, 1k-step warmup, no weight decay), batch size 7, and early stopping. For MoEs, a load-balancing loss with weight 0.2 is added. Inference uses beam search (beam 4, length penalty 0.8, repetition penalty 1.3, max 200 tokens).

\section{Experimental Results and Analysis}

Table~\ref{tab:result_table} presents a comprehensive comparison of our proposed SMEAR-MoE projector against the baseline models across four Indian languages on four distinct test sets. The results clearly demonstrate that our SMEAR-MoE architecture achieves the best overall performance, consistently outperforming the hard-gating MoE models and other baselines across nearly all conditions.

\begin{figure}[!h]  
    \label{fig:heatmap}
    \centering
    \begin{subfigure}[b]{0.5\columnwidth}
        \centering
        \includegraphics[width=\linewidth]{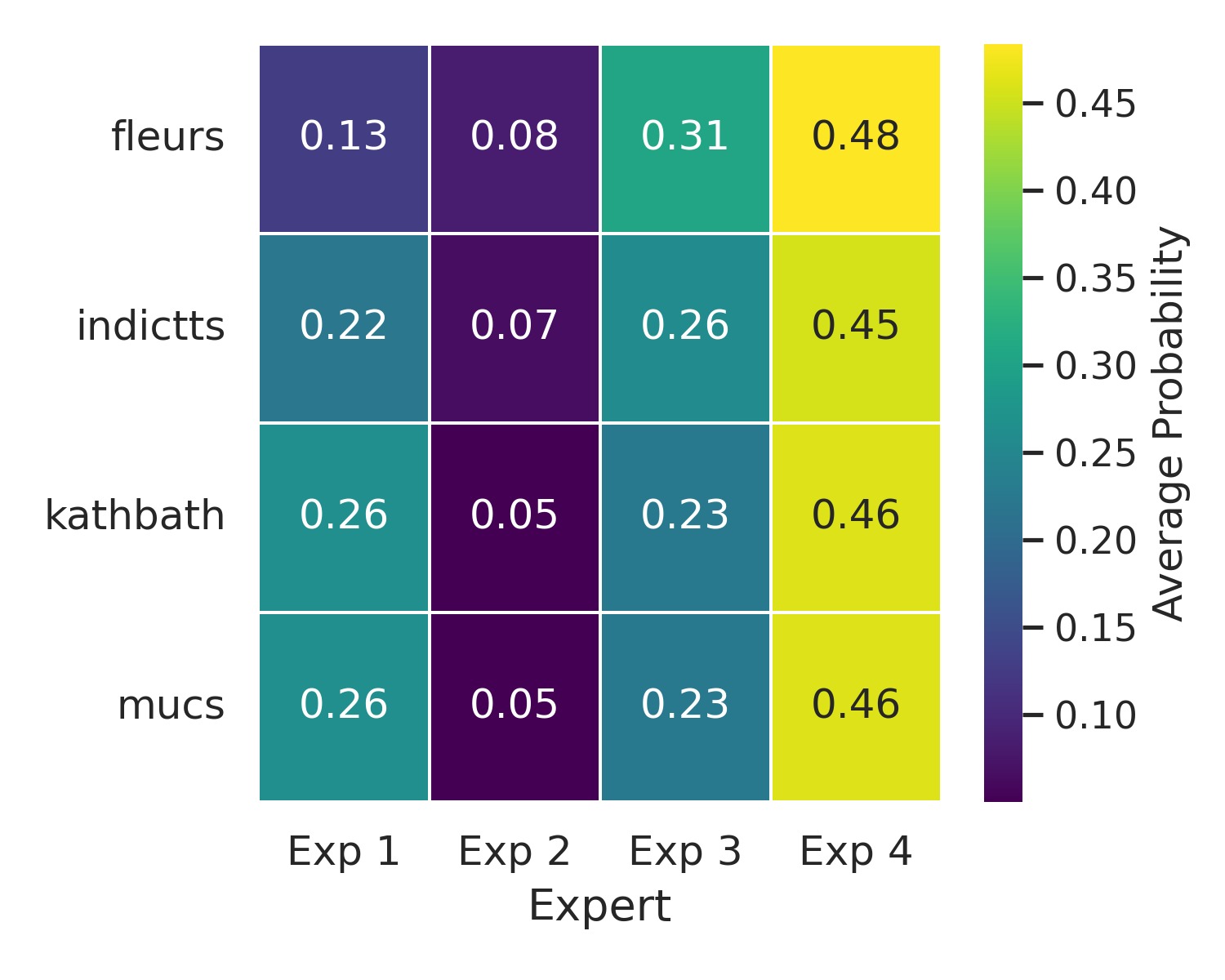}
        \caption{Hindi}
        \label{fig:hi}
    \end{subfigure}
    \hspace{-0.25cm}
    \begin{subfigure}[b]{0.5\columnwidth}
        \centering
        \includegraphics[width=\linewidth]{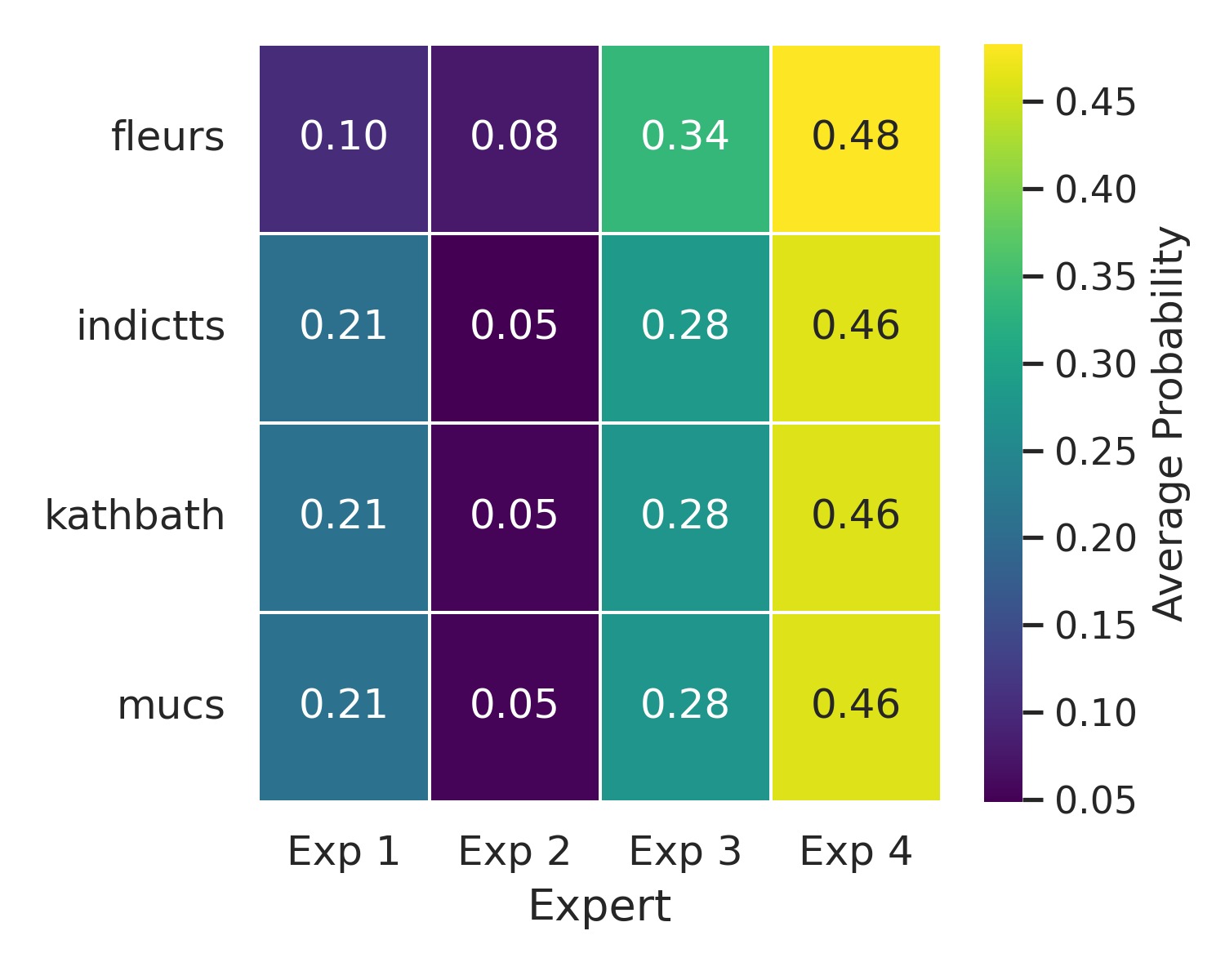}
        \caption{Marathi}
        \label{fig:mr}
    \end{subfigure}
    
    \vspace{0.1em}
    
    \begin{subfigure}[b]{0.5\columnwidth}
        \centering
        \includegraphics[width=\linewidth]{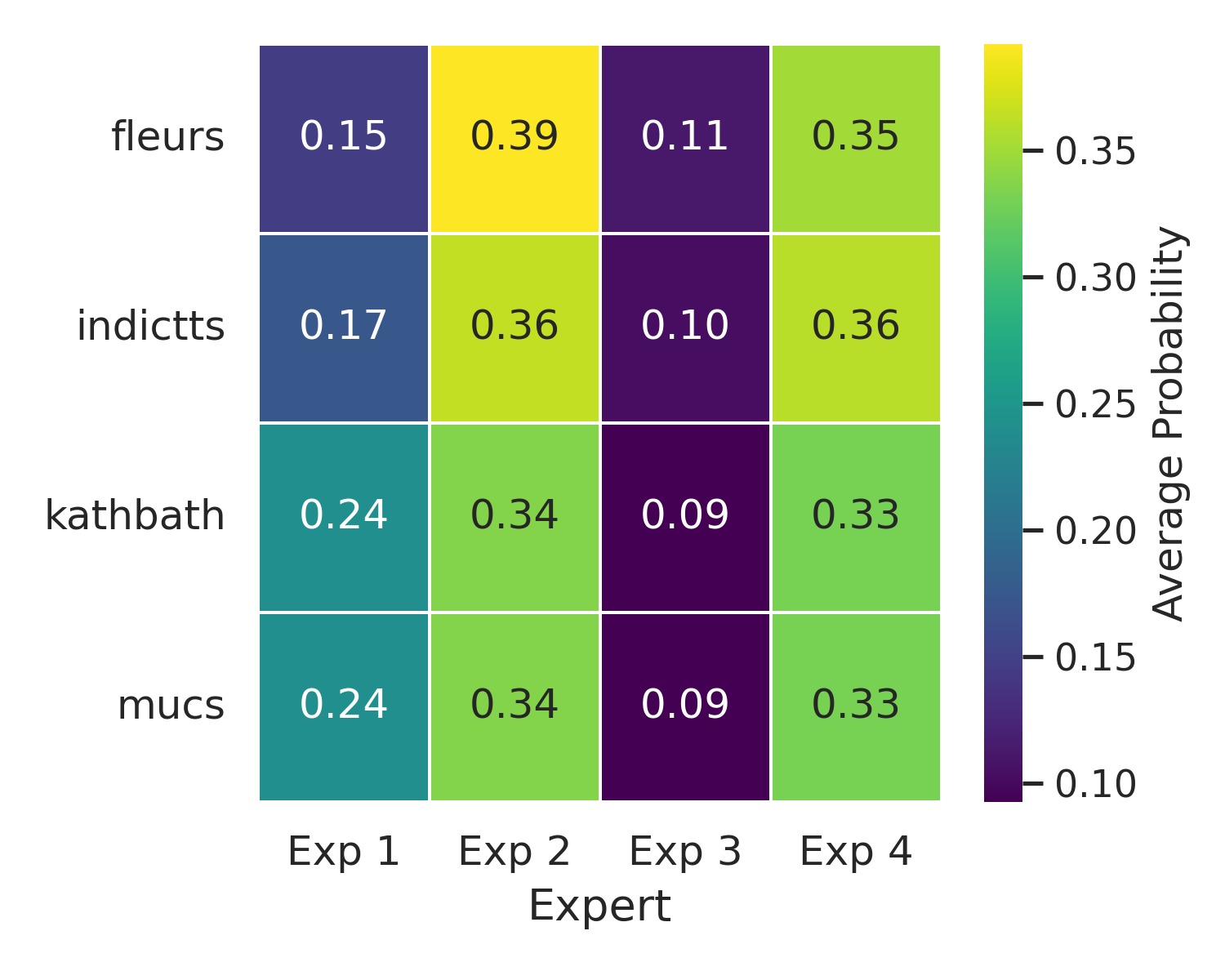}
        \caption{Tamil}
        \label{fig:ta}
    \end{subfigure}
    \hspace{-0.25cm}
    \begin{subfigure}[b]{0.5\columnwidth}
        \centering
        \includegraphics[width=\linewidth]{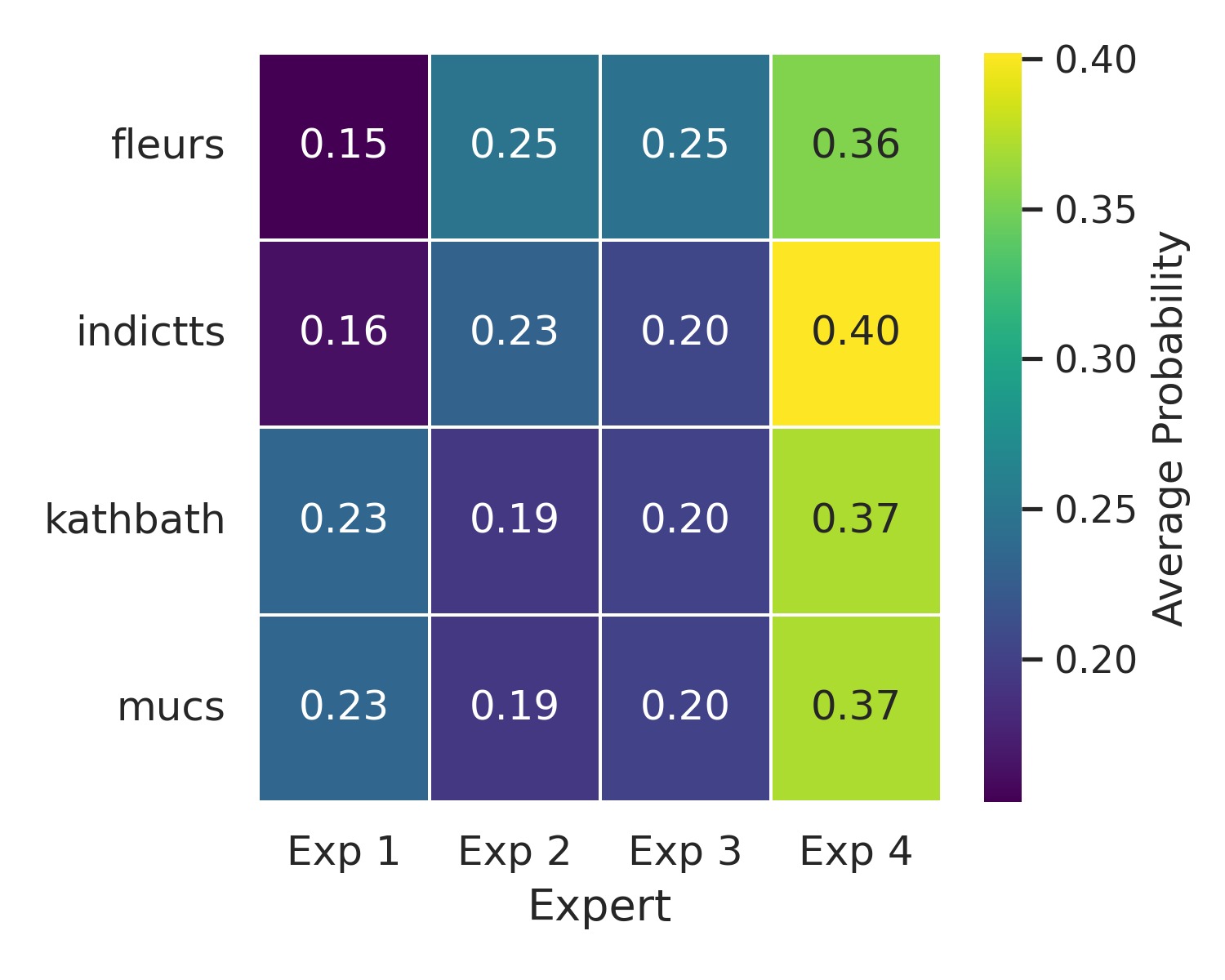}
        \caption{Telugu}
        \label{fig:te}
    \end{subfigure}
    
    \caption{Routing probability heatmaps under SMEAR-MoE, showing meaningful expert specialization: Hindi and Marathi share a dominant expert, Tamil uses a distinct one, while Telugu exhibits a more distributed pattern.}
    \label{fig:all_heatmaps}
    \vspace{-0.5cm}
\end{figure}

\subsection{Performance Comparison}
Our proposed \textbf{SMEAR-MoE} achieves the lowest average WER across all benchmarks, establishing it as the most effective architecture. The \textbf{Dense Ensemble} is consistently the second-best model, confirming that broader parameter access and knowledge sharing are beneficial. However, its static uniform weighting limits cross-lingual adaptation, and its high computational cost reduces practicality. In contrast, the \textbf{Token-Level} and \textbf{Utterance-Level MoE} models underperform due to poor gradient estimation in top-$k$ routing, which the load-balancing loss only partially alleviates, leading to unstable training and under-utilized experts. Finally, we compare SMEAR-MoE against strong baselines, including Whisper Large-v3~\cite{whisper} and SLAM’s single-projector~\cite{ma2024embarrassingly}. 

\subsection{Analysis of Learned Routing Behavior}
Analysis of the routing probabilities from our SMEAR-MoE model (Figures \ref{fig:hi}, \ref{fig:mr}, \ref{fig:ta}, and \ref{fig:te}) reveals that it learns meaningful linguistic specializations without explicit instruction. A shared routing preference is evident for the Indo-Aryan languages \textbf{Hindi} and \textbf{Marathi}, which not only belong to the same family but also share the Devanagari script; both predominantly favor the same expert (e.g., Expert 4), indicating that the model exploits their strong acoustic and structural commonalities. In contrast, \textbf{Tamil} is consistently routed to a different, specialized expert (e.g., Expert 2). The router’s behavior on \textbf{Telugu}, another Dravidian language, is more nuanced: although it shares a family with Tamil, its distinct script and phonological patterns lead to more distributed probabilities, reflecting its divergence. These results demonstrate that SMEAR-MoE not only achieves superior ASR performance but also learns an interpretable routing strategy that mirrors the underlying linguistic relationships in the data, making it a promising approach for multilingual settings.

\xhdr{Runtime Complexity}
We evaluated computational efficiency using Real-Time Factor (RTF) on an NVIDIA H200 GPU. \textbf{SMEAR-MoE} achieves an RTF of 0.198, nearly identical to the single projector baseline (0.196). In contrast, the Dense Ensemble is slower (0.243) due to its higher parameter count. These results show that SMEAR-MoE uniquely combines strong ASR performance with efficiency comparable to a simple monolithic projector.

\section{Conclusion}
We presented SMEAR-MoE, a stabilized Mixture-of-Experts projector for multilingual ASR that overcomes the instability and expert collapse of conventional MoEs. By merging expert parameters through soft gating, our method ensures dense gradient flow, enabling both specialization and cross-lingual sharing. Experiments on four Indic languages show that SMEAR-MoE significantly outperforms monolithic and static projector baselines while remaining computationally efficient. Routing analysis further revealed clear and interpretable expert specialization aligned with linguistic families, establishing stabilized multi-expert projectors as a promising new direction for scalable, efficient, and robust LLM-based multilingual ASR.

\section{Acknowledgments}
This work is supported by BharatGen\footnote{https://bharatgen.com/}, an Indian Government-funded initiative focused on developing multimodal large language models for Indian languages.

\bibliographystyle{IEEEbib}
\bibliography{strings,refs}

\end{document}